\documentclass[11pt]{article}

\usepackage[preprint]{acl}

\usepackage{times}
\usepackage{latexsym}
\usepackage{multirow}
\usepackage{enumitem} % For custom list settings
\usepackage[T1]{fontenc}
\usepackage[utf8]{inputenc}

\usepackage{microtype}

\usepackage{inconsolata}

\usepackage{graphicx}

\title{CogMem: A Cognitive Memory Architecture for Sustained Multi-Turn Reasoning in Large Language Models}

\author{Yiran Zhang\textsuperscript{1}, Jincheng Hu\textsuperscript{2}, Mark Dras\textsuperscript{1}, \textbf{Usman Naseem}\textsuperscript{1} \\
 Macquarie University\textsuperscript{1},  Independent Researcher\textsuperscript{2} \\
\texttt{\{yiran.zhang, mark.dras, usman.naseem\}@mq.edu.au}}

\begin{document}
\maketitle
\begin{abstract}

Large language models (LLMs) excel at single-turn reasoning but often lose accuracy and coherence over extended, multi-turn interactions. Recent evaluations such as TurnBench highlight recurring failure modes—reasoning bias, task drift, hallucination, overconfidence, and memory decay. Current approaches typically append full conversational histories, causing unbounded context growth, higher computational costs, and degraded reasoning efficiency. We introduce CogMem, a cognitively inspired, memory-augmented LLM architecture that supports sustained iterative reasoning through structured, persistent memory. CogMem incorporates three layers: a Long-Term Memory (LTM) that consolidates cross-session reasoning strategies; a Direct Access (DA) memory that maintains session-level notes and retrieves relevant long-term memories; and a Focus of Attention (FoA) mechanism that dynamically reconstructs concise, task-relevant context at each turn. Experiments on TurnBench show that this layered design mitigates reasoning failures, controls context growth, and improves consistency across extended reasoning chains, moving toward more reliable, human-like reasoning in LLMs.

% Large language models (LLMs) demonstrate strong single-turn reasoning but often struggle to sustain accuracy and coherence across extended multi-turn, multi-step interactions. Recent evaluations such as TurnBench reveal recurring failure patterns, including reasoning bias, task misconception, information hallucination, overconfident inference, and long-term memory decay. In addition to these reasoning failures, most existing systems handle multi-turn reasoning by continuously appending prior context, which leads to unbounded context growth, higher computational cost, and degraded reasoning efficiency over long dialogues. To address these challenges, we present CogMem, a cognitively inspired, memory-augmented LLM architecture that enhances iterative reasoning through structured, persistent memory. Grounded in \citet{oberauer_access_2002}'s tripartite model of working memory, the system integrates three layers: a Long-Term Memory (LTM) that accumulates cross-session strategies and distilled reasoning patterns; a Direct Access (DA) memory that maintains session-level notes of plans and conclusions while activating relevant long-term memories to assist ongoing reasoning; and a Focus of Attention (FoA) mechanism that dynamically reconstructs context at each turn to preserve essential information within token limits. The experiments on TurnBench suggest that this layered memory design mitigates reasoning failures, controls context growth, and improves consistency over extended reasoning chains, offering a promising path toward more reliable, human-like LLM reasoning.
\end{abstract}

\section{Introduction}

Reasoning is central to human intelligence~\citep{Wason1972-WASPOR-3, 10.1093/oxfordhb/9780199734689.001.0001} and serves as a core benchmark for artificial systems. Large language models (LLMs) have demonstrated remarkable performance on single-turn reasoning tasks~\citep{10.5555/3495724.3495883, touvron2023llamaopenefficientfoundation, comanici2025gemini25pushingfrontier}. Yet real-world applications demand iterative, multi-turn reasoning: models must integrate new information, revise assumptions, and maintain coherent thinking over extended interactions. Sustaining long-horizon reasoning remains a key challenge.

Recent benchmarks have highlighted this gap. TurnBench-MS~\citep{zhang2025turnbenchmsbenchmarkevaluatingmultiturn} requires models to infer hidden rules through sequential interactions, integrating feedback across turns. Even top-performing LLMs achieve only 84\% accuracy in classic mode and 18\% in nightmare mode, with early errors often cascading through subsequent steps. Other benchmarks, including AvalonBench~\citep{light2023avalonbench} and MT-Bench~\citep{bai2024mt}, confirm that sustained dialog-based reasoning remains a significant challenge. Chain-of-Thought prompting~\citep{wei2022chain} improves single-turn inference but is insufficient for long-term, multi-turn reasoning, highlighting the need for persistent, structured memory.

\begin{figure*}[!t]
  \centering
  \includegraphics[width=.85\textwidth]{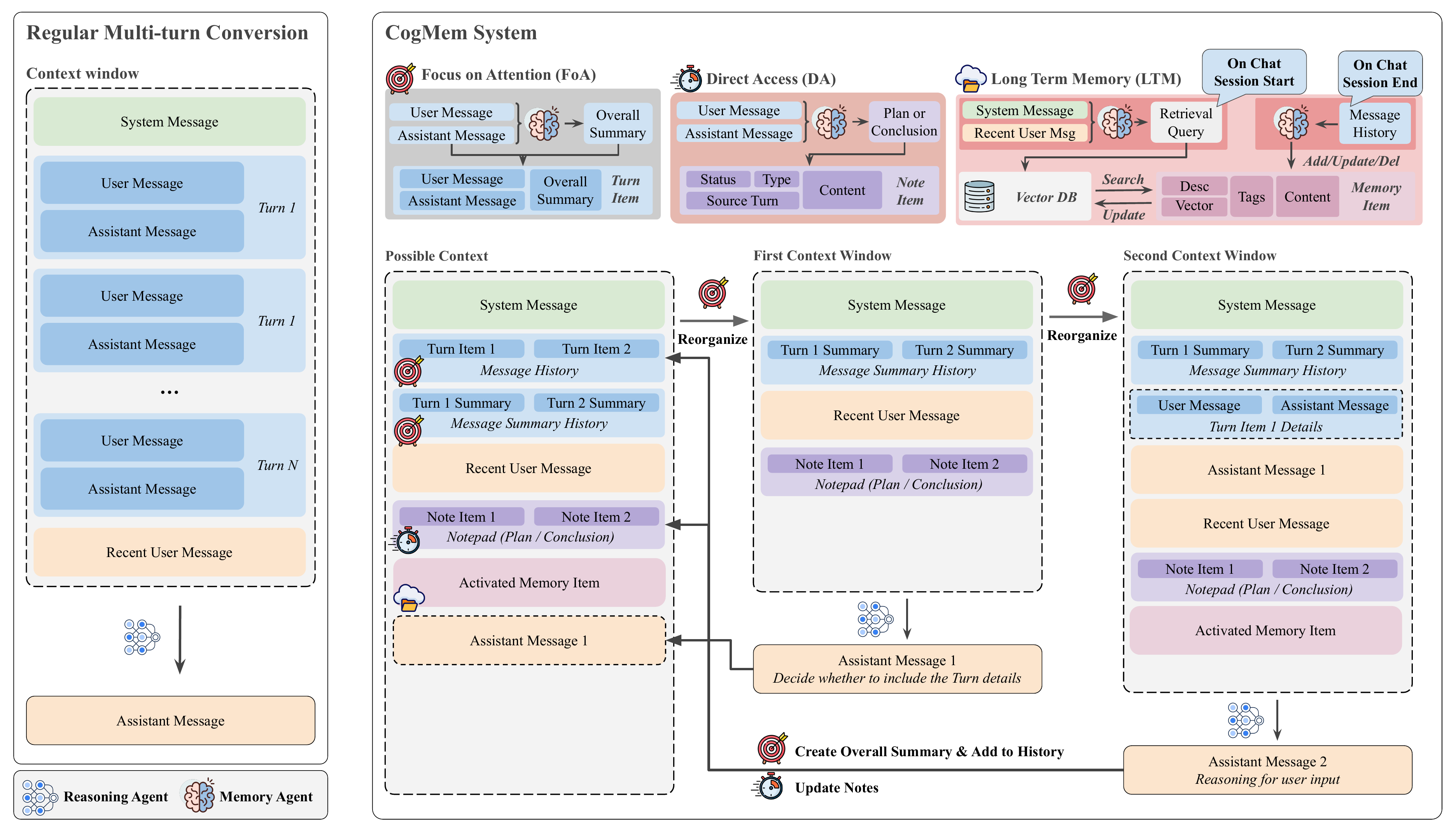}
  \caption{Overview of the CogMem framework.}
  \label{fig:overall}
\end{figure*}

Memory-augmented LLMs offer one path forward. Retrieval-Augmented Generation~\citep{lewis2020retrieval} extends context via retrieval but lacks persistent lifecycle management. Later systems, including MemBank~\citep{zhong2024memorybank}, Mem0~\citep{chhikara2025mem0}, Mem1~\citep{zhou2025mem1}, A-Mem~\citep{xu2025mem}, and MemOS~\citep{li2025memos}, introduced structured, evolving memory architectures that enable dynamic extraction, updating, and long-term storage of relevant knowledge. However, these approaches typically treat memory and reasoning as loosely coupled.

We propose a cognitively inspired, memory-augmented reasoning architecture that tightly integrates reasoning and structured memory. The system maintains a hierarchical memory: a long-term memory stores distilled reasoning strategies and cross-session knowledge, a direct-access memory preserves session-relevant plans and information, and a focus-of-attention module reconstructs minimal working context at each turn. Reasoning and memory agents collaborate continuously, allowing models to recover from early errors, maintain coherence, and efficiently manage context across sessions. This framework combines insights from prior memory-augmented LLMs with a unified, cognitively grounded approach to persistent, adaptive reasoning. Our contributions are:

% \paragraph{Our contributions are:}  
\begin{itemize}[noitemsep,leftmargin=*]
    \item A structured, multi-level memory architecture that enables coherent, long-horizon reasoning while controlling context growth.  
    \item Integration of modular reasoning and memory agents for continuous inference, knowledge accumulation, and recovery from reasoning errors.  
    \item Empirical validation on TurnBench-MS demonstrating improved reasoning accuracy, reduced failure modes, and enhanced efficiency in extended multi-turn reasoning tasks.
\end{itemize}

\section{Method}

\subsection{System Overview}

We propose a memory-augmented reasoning framework designed to enable LLMs to maintain coherent, adaptive, and efficient reasoning across multiple turns. The system comprises two cooperative agents: \textbf{the reasoning agent} and \textbf{the memory agent}, coordinated by a \textbf{session manager} and a \textbf{memory manager} (Section~\ref{sec:implementation_efficiency}). These components operate on a hierarchical memory structure inspired by cognitive models of working memory, consisting of three layers with distinct levels of accessibility and persistence: long-term memory (LTM), direct-access memory (DA), and focus of attention (FoA) (Figure~\ref{fig:overall}).

\textbf{Long-Term Memory (LTM)} stores distilled reasoning strategies and reusable problem-solving patterns accumulated across sessions, serving as a persistent repository that evolves as new insights are integrated.

\textbf{Direct-Access Memory (DA)} functions as session-level working memory, maintaining concise notes of intermediate conclusions, sub-goals, and ongoing plans. At the start of each session, the LTM supplies the DA with the most relevant long-term memories, enabling the reasoning agent to access prior knowledge without repeatedly querying the full LTM.

\textbf{Focus of Attention (FoA)} dynamically reconstructs the minimal reasoning context for each turn. It selectively integrates current notes, retrieved long-term memories, summarized dialogue history, and new user input to produce a compact, informative prompt. This ensures a bounded, interpretable context while retaining essential information for accurate reasoning.

This hierarchical design allows iterative reasoning without unbounded context expansion, preserving critical reasoning cues, recovering from earlier mistakes, and maintaining continuity across extended reasoning chains while controlling computational cost.

\subsection{Processing Workflow}

Upon receiving user input, the session manager determines whether an existing session can be reused or extended. If a matching session exists, stored results and memory states are reused directly. Otherwise, the system searches for an \emph{inheritable session}, where a prior dialogue forms the prefix of the current input. In these cases, corresponding memories and notes are restored to continue reasoning seamlessly. If no session is reusable, a new session is initialized: the DA is created empty, and the memory agent queries the LTM for relevant information by rewriting the current situation into a descriptive query for semantic retrieval.

Reasoning begins with constructing the first FoA, the \emph{First Context Window} (Figure~\ref{fig:overall}b), which includes session notes, summarized dialogue history with turn identifiers, and new user input. The reasoning agent evaluates whether this context is sufficient; if incomplete, it returns identifiers of missing turns, allowing FoA to retrieve details and construct a \emph{Second Context Window}. The agent then performs inference over this refined context, integrating retrieved turn details, user input, and existing notes. Together, these cycles constitute a single turn in the dialogue flow.

After reasoning, the memory agent summarizes the generated content. The summary, user input, and model response form a new turn record, updating the DA asynchronously. This ensures the user receives immediate responses while the memory remains coherent and current. Turn records and updated memory states are stored in the session cache for efficient reuse.

Inactive sessions trigger a cleanup procedure: expired sessions and cached DA entries are deleted, while FoA turn data are reference-counted to enable turn-level reuse. Before final removal, the LTM reviews the session to determine whether distilled knowledge should be merged, added, or excluded, ensuring only meaningful reasoning traces persist.

\subsection{Interaction among Memory Layers}

The memory layers interact through ongoing cycles of retrieval, reconstruction, and refinement. FoA selects activated memories and relevant DA notes to construct the next reasoning context. The DA retrieves relevant entries from the LTM and updates its notes as new reasoning outcomes are summarized. At session conclusion, the LTM distills essential insights from DA and FoA summaries, integrating them into its persistent store. This bidirectional flow allows the model to accumulate experience and maintain stable reasoning across sessions.

\subsection{Implementation and Efficiency}
\label{sec:implementation_efficiency}

The framework is modular and compatible with the OpenAI SDK, operating with any base LLM. The memory agent is a lightweight model responsible for summarization, retrieval, and note generation, enabling a separation of complex inference and efficient memory management.

Short-term data, including notes, turns, and session metadata, are stored in RAM for fast access, optionally supported by external caches such as Redis. The LTM resides in a vectorized database for high-speed semantic retrieval and selective updates. The memory manager orchestrates interactions among FoA, DA, and LTM, while the session manager handles identification, caching, expiration, and garbage collection. Event-triggered garbage collection ensures consistent latency regardless of session length, maintaining bounded computational cost across long dialogues.

\subsection{Design Rationale and Error Mitigation}

The framework addresses recurring multi-turn reasoning failures, including reasoning bias, task misconception, hallucination, overconfidence, memory decay, and unbounded context growth.

FoA and DA reduce reasoning bias, hallucination, and memory decay by reconstructing a concise input window at each step, retaining only essential reasoning traces. DA maintains structured notes of current plans and intermediate conclusions rather than full histories, ensuring reasoning is grounded in distilled information. Iteratively revising assumptions in notes reduces bias over time.

LTM accumulates refined reasoning patterns and task-specific strategies across sessions, guiding task understanding and planning, and enhancing cross-session adaptability. Persistent memory supports incremental improvement and long-term stability.

The session manager and lifecycle controller bound computational cost. Reusable or inheritable sessions avoid redundant computation, and selective refinement into LTM ensures only meaningful reasoning traces persist. Overall, the framework transforms multi-turn reasoning from an ever-growing sequence of appended contexts into structured, iterative refinement, yielding more reliable, interpretable, and temporally consistent reasoning.

\begin{table}[ht]
\centering
\resizebox{\linewidth}{!}{%
\begin{tabular}{lccccc}
\hline
\multicolumn{1}{c}{\multirow{2}{*}{Models}} &  & \multicolumn{4}{c}{Average Accuracy} \\ \cline{3-6} 
\multicolumn{1}{c}{} &  & Total & Easy & Medium & Hard \\ \hline
Baseline - gemini-2.5-flash &  & 0.76 & 0.87 & 0.93 & 0.47 \\ \hline
Baseline + FoA &  & 0.76 & 0.93 & 0.84 & 0.53 \\
Baseline + FoA + DA &  & 0.84 & 0.93 & 0.93 & 0.66 \\
Baseline + FoA + DA + LTM & \multicolumn{1}{l}{} & \textbf{0.93} & \textbf{1.00} & \textbf{1.00} & \textbf{0.80} \\ \hline
Random Guess &  & 0.0085 & 0.0079 & 0.0098 & 0.0077 \\ \hline
\end{tabular}
}
\caption{Performance comparison between the CogMem framework and the baseline model under the TurnBench classic mode settings. A random-guessing baseline is included for comparison. All results are obtained using the chain-of-thought (CoT) prompting strategy. The \textbf{bold} text represents the best performance.}
\label{tab:result_table}
\end{table}

\section{Experiment Setup and Results}

\subsection{Experimental Settings}

% We evaluate our framework on the TurnBench-MS benchmark~\citep{zhang2025turnbenchmsbenchmarkevaluatingmultiturn}, designed to test iterative reasoning and adaptation in multi-turn interactions. TurnBench simulates rule-discovery games where models infer hidden patterns through sequential queries and feedback. Two evaluation modes, \emph{classic} and \emph{nightmare}, capture different levels of uncertainty and contextual complexity. Following the official protocol, we use 45 setups per mode across three difficulty levels (easy, medium, hard), totaling 90 games. 
We evaluate our framework on the TurnBench-MS benchmark~\citep{zhang2025turnbenchmsbenchmarkevaluatingmultiturn}, which is designed to assess iterative reasoning and adaptation in multi-turn interactions. TurnBench simulates rule-discovery games in which models must infer hidden patterns through sequential queries and feedback. Following the official evaluation protocol, we use the classic mode consisting of 45 game setups across three difficulty levels, including easy, medium, and hard.

All experiments employ chain-of-thought prompting. The reasoning agent is Gemini 2.5 Flash~\citep{comanici2025gemini25pushingfrontier}, while the memory agent uses a lightweight Gemini 2.5 Flash Lite variant for summarization and retrieval. External vector storage is managed via Milvus~\citep{wang2021milvus}, and accuracy is computed following TurnBench standards~\citep{pedregosa2011scikit}. Full experimental and implementation details are provided in Appendix~\ref{sec:appendix_experiments}.

\subsection{Baselines}

We evaluate four incremental configurations to isolate the contribution of each memory component:  

\begin{itemize}[noitemsep,leftmargin=*]
    \item \textbf{Baseline:} Gemini 2.5 Flash with chain-of-thought prompting, no memory.  
    \item \textbf{Baseline + FoA:} Adds focus-of-attention for bounded context reconstruction.  
    \item \textbf{Baseline + FoA + DA:} Incorporates direct-access memory for structured session notes.  
    \item \textbf{Baseline + FoA + DA + LTM:} Full system with long-term memory for cross-session knowledge accumulation.  
\end{itemize}

All other settings (prompts, decoding parameters, temperature) are identical. A random-guessing baseline is included to indicate lower-bound performance. Detailed experiment setups, prompts, and memory configurations are in Appendix~\ref{sec:appendix_experiments}.

\subsection{Results and Analysis}
Table \ref{tab:result_table} summarizes the performance across the classic mode of TurnBench-MS. The baseline uses the official Gemini 2.5 Flash results, achieving an overall accuracy of 0.76. Introducing the FoA alone brings a small but consistent improvement on harder setups in the classic environment, suggesting that dynamic context reconstruction helps reduce redundant reasoning and maintain focus, though its impact remains limited under high-noise conditions.

Adding the direct-access memory yields a substantial gain, raising total accuracy in the classic mode to 0.84. This indicates that structured note keeping and short-term summarization improve reasoning consistency and help the model recover from earlier mistakes by preserving key conclusions between turns. Finally, integrating the long-term memory produces the strongest results: 0.93 total accuracy in the classic mode, with perfect scores on easy and medium difficulties and 0.80 on hard setups. The improvement confirms that cross-session accumulation of distilled reasoning patterns allows the model to generalize strategies and stabilize its decision process over extended reasoning chains.

Overall, the ablation trend demonstrates clear additive benefits from each memory layer. FoA mitigates token growth and focus drift; DA enhances within-session consistency; and LTM extends adaptation across sessions. Together, they transform the model’s behavior from reactive, context-bound reasoning to structured, self-referential reasoning capable of maintaining coherence under both standard and adversarial conditions.

\section{Conclusion}
We introduced CogMem, a cognitively inspired memory architecture that enables large language models to sustain accurate, coherent, and efficient reasoning over extended multi-turn interactions. CogMem integrates three complementary layers—long-term memory, direct-access memory, and focus of attention—to manage reasoning information at different temporal scales. This design allows the model to retain essential context, prevent redundant expansion, and recover from earlier errors. Experiments on TurnBench-MS show that each memory layer contributes incrementally to performance, with the full CogMem system achieving substantial gains over the baseline Gemini 2.5 Flash model. These results demonstrate that structured, cognitively grounded memory can transform LLMs from reactive, context-dependent reasoners into adaptive, self-consistent systems capable of sustained multi-turn inference.

\clearpage

\section*{Limitations}

While CogMem demonstrates clear improvements in reasoning stability and coherence, its current evaluation remains limited in scope. All experiments were conducted on the TurnBench-MS benchmark with a single base model, which constrains the generalizability of the findings. Future work will extend the evaluation to a wider range of reasoning datasets, task types, and language models to further validate CogMem’s adaptability and robustness across diverse multi-turn reasoning scenarios.

\section*{Ethical}

This work focuses on improving the reasoning stability and efficiency of large language models through structured memory design. All experiments were conducted using publicly available benchmark data (TurnBench-MS) and proprietary language models accessed through standard APIs. No personally identifiable or sensitive information was used in any stage of development or evaluation. The goal of CogMem is to enhance the reliability and interpretability of multi-turn reasoning systems, which may reduce unintended hallucinations and overconfident outputs in deployed AI applications. Nonetheless, improved long-term reasoning capabilities also raise considerations regarding information persistence and user privacy. Future work will further examine safeguards for memory management, data retention, and user consent to ensure that cognitively inspired memory systems remain transparent, controllable, and aligned with ethical standards for responsible AI development.

\bibliography{custom}

@article{oberauer_access_2002,
	title = {Access to information in working memory: {Exploring} the focus of attention.},
	volume = {28},
	issn = {1939-1285, 0278-7393},
	shorttitle = {Access to information in working memory},
	url = {https://doi.apa.org/doi/10.1037/0278-7393.28.3.411},
	doi = {10.1037/0278-7393.28.3.411},
	language = {en},
	number = {3},
	urldate = {2025-10-06},
	journal = {Journal of Experimental Psychology: Learning, Memory, and Cognition},
	author = {Oberauer, Klaus},
	year = {2002},
	pages = {411--421},
}

@inproceedings{10.5555/3495724.3495883,
author = {Brown, Tom B. and Mann, Benjamin and Ryder, Nick and Subbiah, Melanie and Kaplan, Jared and Dhariwal, Prafulla and Neelakantan, Arvind and Shyam, Pranav and Sastry, Girish and Askell, Amanda and Agarwal, Sandhini and Herbert-Voss, Ariel and Krueger, Gretchen and Henighan, Tom and Child, Rewon and Ramesh, Aditya and Ziegler, Daniel M. and Wu, Jeffrey and Winter, Clemens and Hesse, Christopher and Chen, Mark and Sigler, Eric and Litwin, Mateusz and Gray, Scott and Chess, Benjamin and Clark, Jack and Berner, Christopher and McCandlish, Sam and Radford, Alec and Sutskever, Ilya and Amodei, Dario},
title = {Language models are few-shot learners},
year = {2020},
isbn = {9781713829546},
publisher = {Curran Associates Inc.},
address = {Red Hook, NY, USA},
booktitle = {Proceedings of the 34th International Conference on Neural Information Processing Systems},
articleno = {159},
numpages = {25},
location = {Vancouver, BC, Canada},
series = {NIPS '20}
}

@misc{touvron2023llamaopenefficientfoundation,
      title={LLaMA: Open and Efficient Foundation Language Models}, 
      author={Hugo Touvron and Thibaut Lavril and Gautier Izacard and Xavier Martinet and Marie-Anne Lachaux and Timothée Lacroix and Baptiste Rozière and Naman Goyal and Eric Hambro and Faisal Azhar and Aurelien Rodriguez and Armand Joulin and Edouard Grave and Guillaume Lample},
      year={2023},
      eprint={2302.13971},
      archivePrefix={arXiv},
      primaryClass={cs.CL},
      url={https://arxiv.org/abs/2302.13971}, 
}

@article{comanici2025gemini25pushingfrontier,
  title={Gemini 2.5: Pushing the frontier with advanced reasoning, multimodality, long context, and next generation agentic capabilities},
  author={Comanici, Gheorghe and Bieber, Eric and Schaekermann, Mike and Pasupat, Ice and Sachdeva, Noveen and Dhillon, Inderjit and Blistein, Marcel and Ram, Ori and Zhang, Dan and Rosen, Evan and others},
  journal={arXiv preprint arXiv:2507.06261},
  year={2025}
}

@book{Wason1972-WASPOR-3,
	address = {Cambridge, MA, USA},
	author = {Peter Cathcart Wason and Philip Nicholas Johnson{-}Laird},
	editor = {},
	publisher = {Harvard University Press},
	title = {Psychology of Reasoning: Structure and Content},
	year = {1972}
}

@book{10.1093/oxfordhb/9780199734689.001.0001,
    author = {Holyoak, Keith J. and Morrison, Robert G.},
    title = {The Oxford Handbook of Thinking and Reasoning},
    publisher = {Oxford University Press},
    year = {2012},
    month = {03},
    isbn = {9780199734689},
    doi = {10.1093/oxfordhb/9780199734689.001.0001},
    url = {https://doi.org/10.1093/oxfordhb/9780199734689.001.0001},
}

@inproceedings{zhang2025turnbenchmsbenchmarkevaluatingmultiturn,
    title = "{T}urn{B}ench-{MS}: A Benchmark for Evaluating Multi-Turn, Multi-Step Reasoning in Large Language Models",
    author = "Zhang, Yiran  and
      Wang, Mo  and
      Li, Xiaoyang  and
      Ren, Kaixuan  and
      Zhu, Chencheng  and
      Naseem, Usman",
    editor = "Christodoulopoulos, Christos  and
      Chakraborty, Tanmoy  and
      Rose, Carolyn  and
      Peng, Violet",
    booktitle = "Findings of the Association for Computational Linguistics: EMNLP 2025",
    month = nov,
    year = "2025",
    address = "Suzhou, China",
    publisher = "Association for Computational Linguistics",
    url = "https://aclanthology.org/2025.findings-emnlp.1084/",
    doi = "10.18653/v1/2025.findings-emnlp.1084",
    pages = "19892--19924",
    ISBN = "979-8-89176-335-7"
}

@article{wei2022chain,
  title={Chain-of-thought prompting elicits reasoning in large language models},
  author={Wei, Jason and Wang, Xuezhi and Schuurmans, Dale and Bosma, Maarten and Xia, Fei and Chi, Ed and Le, Quoc V and Zhou, Denny and others},
  journal={Advances in neural information processing systems},
  volume={35},
  pages={24824--24837},
  year={2022}
}

@article{zhou2025mem1,
  title={MEM1: Learning to Synergize Memory and Reasoning for Efficient Long-Horizon Agents},
  author={Zhou, Zijian and Qu, Ao and Wu, Zhaoxuan and Kim, Sunghwan and Prakash, Alok and Rus, Daniela and Zhao, Jinhua and Low, Bryan Kian Hsiang and Liang, Paul Pu},
  journal={arXiv preprint arXiv:2506.15841},
  year={2025}
}

@article{light2023avalonbench,
  title={Avalonbench: Evaluating llms playing the game of avalon},
  author={Light, Jonathan and Cai, Min and Shen, Sheng and Hu, Ziniu},
  journal={arXiv preprint arXiv:2310.05036},
  year={2023}
}

@article{bai2024mt,
  title={Mt-bench-101: A fine-grained benchmark for evaluating large language models in multi-turn dialogues},
  author={Bai, Ge and Liu, Jie and Bu, Xingyuan and He, Yancheng and Liu, Jiaheng and Zhou, Zhanhui and Lin, Zhuoran and Su, Wenbo and Ge, Tiezheng and Zheng, Bo and others},
  journal={arXiv preprint arXiv:2402.14762},
  year={2024}
}

@article{lewis2020retrieval,
  title={Retrieval-augmented generation for knowledge-intensive nlp tasks},
  author={Lewis, Patrick and Perez, Ethan and Piktus, Aleksandra and Petroni, Fabio and Karpukhin, Vladimir and Goyal, Naman and K{\"u}ttler, Heinrich and Lewis, Mike and Yih, Wen-tau and Rockt{\"a}schel, Tim and others},
  journal={Advances in neural information processing systems},
  volume={33},
  pages={9459--9474},
  year={2020}
}

@inproceedings{zhong2024memorybank,
  title={Memorybank: Enhancing large language models with long-term memory},
  author={Zhong, Wanjun and Guo, Lianghong and Gao, Qiqi and Ye, He and Wang, Yanlin},
  booktitle={Proceedings of the AAAI Conference on Artificial Intelligence},
  volume={38},
  number={17},
  pages={19724--19731},
  year={2024}
}

@article{chhikara2025mem0,
  title={Mem0: Building production-ready ai agents with scalable long-term memory},
  author={Chhikara, Prateek and Khant, Dev and Aryan, Saket and Singh, Taranjeet and Yadav, Deshraj},
  journal={arXiv preprint arXiv:2504.19413},
  year={2025}
}

@article{xu2025mem,
  title={A-mem: Agentic memory for llm agents},
  author={Xu, Wujiang and Mei, Kai and Gao, Hang and Tan, Juntao and Liang, Zujie and Zhang, Yongfeng},
  journal={arXiv preprint arXiv:2502.12110},
  year={2025}
}

@article{li2025memos,
  title={MemOS: An Operating System for Memory-Augmented Generation (MAG) in Large Language Models},
  author={Li, Zhiyu and Song, Shichao and Wang, Hanyu and Niu, Simin and Chen, Ding and Yang, Jiawei and Xi, Chenyang and Lai, Huayi and Zhao, Jihao and Wang, Yezhaohui and others},
  journal={arXiv preprint arXiv:2505.22101},
  year={2025}
}

@inproceedings{wang2021milvus,
  title={Milvus: A purpose-built vector data management system},
  author={Wang, Jianguo and Yi, Xiaomeng and Guo, Rentong and Jin, Hai and Xu, Peng and Li, Shengjun and Wang, Xiangyu and Guo, Xiangzhou and Li, Chengming and Xu, Xiaohai and others},
  booktitle={Proceedings of the 2021 international conference on management of data},
  pages={2614--2627},
  year={2021}
}

@article{pedregosa2011scikit,
  title={Scikit-learn: Machine learning in Python},
  author={Pedregosa, Fabian and Varoquaux, Ga{\"e}l and Gramfort, Alexandre and Michel, Vincent and Thirion, Bertrand and Grisel, Olivier and Blondel, Mathieu and Prettenhofer, Peter and Weiss, Ron and Dubourg, Vincent and others},
  journal={the Journal of machine Learning research},
  volume={12},
  pages={2825--2830},
  year={2011},
  publisher={JMLR. org}
}

\clearpage

\appendix
\section{Experimental Details}
\label{sec:appendix_experiments}

\subsection{Benchmark Setup}

We follow the TurnBench-MS protocol~\citep{zhang2025turnbenchmsbenchmarkevaluatingmultiturn} with 45 setups for classic mode and three difficulty levels (easy, medium, hard). Each setup simulates a rule-discovery game requiring multi-turn reasoning and iterative feedback integration.

\subsection{Model Configuration}

\paragraph{Reasoning Agent} Gemini 2.5 Flash is used with standard chain-of-thought prompts. Parameters match those reported in the TurnBench paper to ensure fair comparison.  

\paragraph{Memory Agent} Gemini 2.5 Flash Lite is used for summarization, retrieval, and note maintenance. Its generative reasoning is disabled to improve speed and reduce cost.  

\paragraph{Vector Storage} All memory retrievals use Milvus~\citep{wang2021milvus}. Session and long-term memories are stored as vectors and retrieved using semantic similarity queries.  

\subsection{Prompting and CoT Setup}

- All models use chain-of-thought prompting.  
- Prompts are consistent across all configurations.  
- Memory components are only active in FoA, DA, and LTM settings, not in the baseline.  

\subsection{Baselines and Ablations}

- \textbf{Baseline:} Gemini 2.5 Flash with CoT, no memory.  
- \textbf{Baseline + FoA:} Adds focus-of-attention for bounded context reconstruction.  
- \textbf{Baseline + FoA + DA:} Adds session-level direct-access memory.  
- \textbf{Baseline + FoA + DA + LTM:} Full memory hierarchy for cross-session knowledge accumulation.  
- \textbf{Random-guessing:} Lower-bound reference.  

\subsection{Evaluation Metrics}

Accuracy is computed using scikit-learn~\citep{pedregosa2011scikit}, comparing model outputs against ground-truth rules in each TurnBench scenario. Scores are reported per difficulty level and evaluation mode.

\subsection{Implementation Notes}

- Short-term notes, turns, and session metadata are stored in RAM with optional external caching (Redis).  
- LTM is stored in a vectorized database supporting selective updates.  
- Reasoning and memory agents run asynchronously; the memory agent updates DA after reasoning steps.  
- Session manager handles identification, caching, expiration, and garbage collection.  

All hyperparameters, session management protocols, and retrieval configurations are fully documented in the internal code repository (link omitted for anonymization in submission).

\section{Related Work}

% \section{Related Work}
\paragraph{Multi-Turn Reasoning and Benchmarks}

As LLMs are applied to increasingly complex tasks, benchmarks have evolved to test multi-turn and multi-step reasoning beyond single-query question answering. TurnBench-MS~\citep{zhang2025turnbenchmsbenchmarkevaluatingmultiturn} is a recent benchmark explicitly designed for iterative reasoning: models must infer hidden rules through sequential interactions and integrate feedback across turns. Results reveal a large performance gap between humans and current LLMs, especially when long reasoning chains or corrections are required. Analyses further show that reasoning errors made early in the process often persist or cascade through later steps. Other benchmarks such as AvalonBench~\citep{light2023avalonbench} and MT-Bench~\citep{bai2024mt} confirm that sustained dialog-based reasoning remains a significant challenge for LLMs.

To mitigate such failures, Chain-of-Thought (CoT) prompting~\citep{wei2022chain} encourages models to articulate intermediate steps, improving single-instance reasoning but remaining insufficient for multi-turn or session-level reasoning, where memory continuity and selective recall are essential. Our work builds upon these insights, using TurnBench as a testbed while introducing an explicit structured memory framework to improve reasoning persistence and efficiency.

\paragraph{Memory-Augmented LLMs and Long-Term Interaction}

Recent research increasingly views memory as a core component of scalable LLM systems. Early approaches such as Retrieval-Augmented Generation (RAG)~\citep{lewis2020retrieval} expanded effective context length by incorporating external retrieval, but lacked persistent storage or lifecycle management. Subsequent systems formalized memory as a managed and evolving resource rather than a static cache.

Among these, MemBank~\citep{zhong2024memorybank} introduced a unified external memory store for continual distillation and adaptive overwriting. Mem0~\citep{chhikara2025mem0} and its graph-based variant extended this paradigm to conversational agents, dynamically extracting, updating, and organizing salient facts from dialogue. Mem1~\citep{zhou2025mem1} and A-Mem~\citep{xu2025mem} further explored agentic and self-organizing memories, enabling models to autonomously create and refine memory entries as they interact. MemOS~\citep{li2025memos} conceptualizes memory as a first-class operating resource, establishing unified governance and lifecycle management across parametric, activation, and external memories

These developments collectively illustrate a transition from context extension to structured, persistent, and self-adaptive memory architectures that support long-term reasoning and continuity across sessions. But they generally treat reasoning and memory as loosely coupled components. In contrast, our approach integrates these elements within a unified, cognitively grounded framework. The memory hierarchy is explicitly inspired by \citet{oberauer_access_2002}'s model, mapping the relationships among focus, working, and long-term memory to guide reasoning dynamics. Through tight interaction between the reasoning and memory agents, the system continuously refines and reconstructs context at each turn, rather than updating memory only after reasoning is complete. Furthermore, we introduce session-level lifecycle management that handles session inheritance, recycling, and selective long-term refinement, thereby enabling reasoning continuity and governance beyond what previous systems support.

\section{Efficiency and Token Usage Analysis}
\label{apx:token_analysis}
\begin{figure}[!t]
  \centering
  \includegraphics[width=\linewidth]{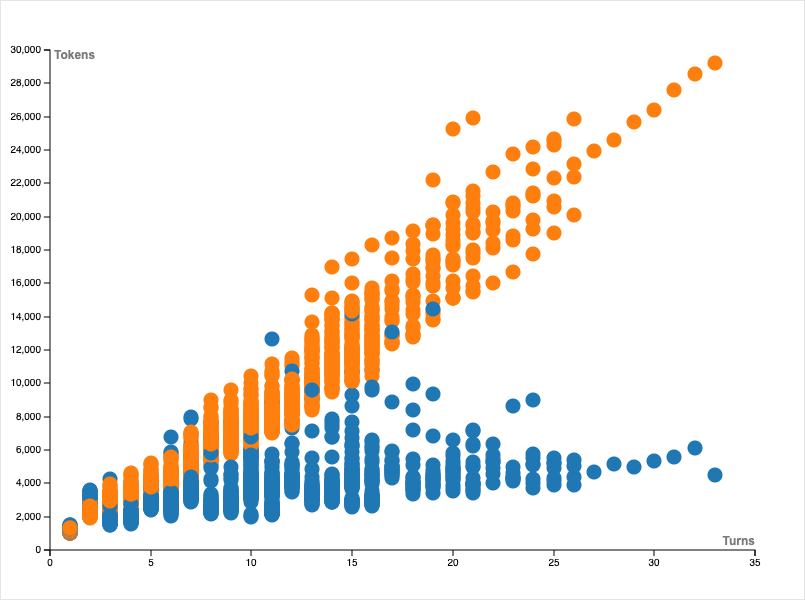}
  \caption{Token consumption per dialogue turn for CogMem (blue) and the model-only baseline (orange) on the TurnBench-MS classic mode. The x-axis represents dialogue turns, while the y-axis shows the number of tokens processed at each turn. CogMem maintains a stable token usage as conversations progress, whereas the model-only baseline exhibits nearly linear growth, reflecting the accumulation of full dialogue context..}
  \label{fig:token_analysis}
\end{figure}

Figure \ref{fig:token_analysis} compares token consumption across dialogue turns between CogMem and the model-only baseline. The baseline shows a near-linear increase in token usage as turn count rises, resulting from the continual accumulation of prior dialogue history in the prompt. In contrast, CogMem maintains a bounded and much flatter growth curve. Its focus-of-attention mechanism reconstructs a concise reasoning context at each turn, and its direct-access memory allows the model to reuse structured notes instead of reloading full historical text. This leads to a substantial reduction in token count—on average, CogMem uses less than half the tokens of the baseline after 15 turns, with the gap widening as dialogues extend. The result confirms that CogMem’s layered memory architecture not only improves reasoning accuracy but also enforces computational efficiency, making multi-turn reasoning more scalable for long or iterative interactions.

\end{document}